\documentclass[conference, a4paper]{IEEEtran}
\IEEEoverridecommandlockouts
\usepackage{cite}
\usepackage{amsmath,amssymb,amsfonts}
\usepackage{algorithmic}
\usepackage{graphicx}
\usepackage{textcomp}
\usepackage{xcolor}
\usepackage{array}
\usepackage{multirow}
\usepackage{booktabs}

\usepackage{subfigure}
\usepackage{subfig}
\usepackage{setspace}
\usepackage[utf8]{inputenc}

\renewcommand{\baselinestretch}{1.02} 
\begin{document}

\title{A Hybrid Method for Training Convolutional Neural Networks}

\author{{Vasco Lopes}\\
\IEEEauthorblockA{\textit{Department of Computer Science} \\
\textit{Universidade da Beira Interior}\\
Covilhã, Portugal \\
vasco.lopes@ubi.pt}
\and
{Paulo Fazendeiro}\\
\IEEEauthorblockA{\textit{Department of Computer Science} \\
\textit{Universidade da Beira Interior}\\ \textit{Instituto de Telecomunicações}\\
Covilhã, Portugal \\
fazendeiro@ubi.pt}
}

\maketitle

\begin{abstract}
Artificial Intelligence algorithms have been steadily increasing in popularity and usage. Deep Learning, allows neural networks to be trained using huge datasets and also removes the need for human extracted features, as it automates the feature learning process. In the hearth of training deep neural networks, such as Convolutional Neural Networks, we find backpropagation, that by computing the gradient of the loss function with respect to the weights of the network for a given input, it allows the weights of the network to be adjusted to better perform in the given task. In this paper, we propose a hybrid method that uses both backpropagation and evolutionary strategies to train Convolutional Neural Networks, where the evolutionary strategies are used to help to avoid local minimas and fine-tune the weights, so that the network achieves higher accuracy results. We show that the proposed hybrid method is capable of improving upon regular training in the task of image classification in CIFAR-10, where a VGG16 model was used and the final test results increased 0.61\%, in average, when compared to using only backpropagation.
   
\end{abstract}


\section{Introduction}
In the recent past, we have witnessed how Artificial Intelligence (AI) has been re-shaping the world, mainly because of deep learning, which removed the need for extensive human hand-crafted features \cite{lecun2015deep}. Deep learning not only provided a way to automate the process of extracting features, but more importantly, it paved the way for democratizing AI, since it allowed people with less knowledge about those technologies to take advantage of them. With the exponential growth of deep learning and the way it can aid developers in reaching new heights, new algorithms to solve both new and existing problems, have surfaced, such as Generative Adversarial Networks \cite{goodfellow2014generative}, which has proved to be an excellent generator of faces \cite{karras2018progressive}, objects \cite{NIPS2016_6194}, styles \cite{CycleGAN2017}, text \cite{yu2017seqgan}, attributes \cite{8718508}, among others \cite{wang2019generative}. Moreover, many traditional AI algorithms have been used in a ``deep" way, which is the case of Convolutional Neural Networks (CNNs) \cite{lecun1998gradient}. Due to the recent advances in hardware, enabling deep networks, with dozens of hidden layers, to be trained in useful time, CNNs became one of the most popular algorithms used to solve problems related with computer vision \cite{SCHMIDHUBER201585}. These have been successfully used in image classification tasks \cite{szegedy2015going, he2016deep, huang2017densely, krizhevsky2012imagenet}, object detection \cite{redmon2016you}, biometrics \cite{8101565, zanlorensi2019ocular}, medical image analysis \cite{shen2017deep}, for text analysis \cite{kim-2014-convolutional, conneau-etal-2017-deep}, among many more applications. The reason for CNNs to be widely used is mainly due to the good results they achieve in virtually all the tasks they have been applied to. 

Even though CNNs achieve good results and remove the need for human extracted features, they still have setbacks, being the most prominent ones: the need for a great deal of data and getting stuck in local minimums while training. Both these situations occur because of the training procedure, which is usually conducted using the backpropagation algorithm \cite{rumelhart1986learning}. Backpropagation is one of the fundamental foundations of neural networks, allowing neural networks to be trained by performing a chain rule with two steps: a forward pass to determine the output of the network, and a backward pass to adjust the network's parameters according to the gradient descent of the error function. However, it is common that this backward pass, searching the global minimum to reduce the error in a multi-dimensional space, gets stuck in local minimums. In this paper we propose a hybrid method, that uses both backpropagation and evolutionary algorithm to train CNNs, to tackle the problem of converging and getting stuck in local minimums. The proposed hybrid training method works in two steps: 1) performing regular training using backpropagation for $n$ epochs; 2) continuing the training procedure using backpropagation, whilst periodically pausing to perform the evolutionary algorithm in the weights of the last layer, in order to avoid local minimums. In short, the proposed method works by training using backpropagation until an initial criterion is met, and afterwards, the training procedure using backpropagation and the evolutionary algorithm is resumed until a convenient stopping criterion is met. Notice that in step 1, $n$, can be selected in various ways, such as: until the validation loss ceases decreasing, fixed in advance (e.g., 15 epochs), and others. In our experiments, we focused on defining $n$ depending on the validation set. The reasons to only evolve the last layer weights are: 1) time constraints, because determining the fitness for every individual in this setting is time-costly; 2) because in some transfer learning \cite{pan2009survey} setups, where the original task is closely related with the new one, only the last layer requires to be fine-tuned, which served as an inspiration. 

In short, the contributions of this work can be summarized as a proposal to improve upon the traditional CNN training, by combining both regular training with backpropagation and evolutionary strategies. More important, is the fact that this proposal can in fact aid the training procedure in order to avoid local minimums, and can be used both as a normal training procedure or to fine-tune a previously trained network. We openly release the code for this proposal\footnote{{https://github.com/VascoLopes/HybridCNNTrain}}, which was developed using PyTorch \cite{paszke2019pytorch}.

The remainder of this work is organized as follows. Section \ref{sec:related} presents the related work. Section \ref{sec:proposed} explains the proposed method in detail, and how the evolutionary strategy is conducted. Section \ref{sec:exp} presents the setup of the experiments, the dataset used and discusses the results obtained. Finally, Section \ref{sec:conc}, provides a conclusion.

\section{Related Work}
\label{sec:related}

Evolutionary algorithms have been extensively used to solve parameter optimization problems \cite{back1991survey, back1996evolutionary}. Tim Salimans \textit{et al.} \cite{salimans2017evolution}, show that evolution strategies can be an alternative to reinforcement learning, in tasks such as humanoid locomotion and Atari games. With focus on solving the same problems, Felipe Petroski Such \textit{et al.} \cite{such2017deep} use evolution strategies to evolve the weights of a deep neural network. This method shares similarities with the one proposed in this paper, but the focus on our work is to complement regular training, they evolve the weights of the entire network, starting from random and using one mutation, requiring significant computational power. In the topic of Neuroevolution, Evolutionary strategies have been extensively used to evolve networks \cite{floreano2008neuroevolution, 1299414,stanley2002evolving}. Recently, evolution strategies have also been used to evolve deeper neural networks \cite{liu2017hierarchical,nsganet}, which is the case of RENAS, that uses evolutionary algorithms and reinforcement learning to perform neural architecture search \cite{chen2019renas}.

CNNs architectures that are known to do well in specific tasks and can easily be transferred to other problems and still perform well are game-changers, which are used in many fields and problems, and often improve the results of the previous state-of-the-art in such tasks, which is the case of image classification, and others \cite{lecun1998gradient, krizhevsky2012imagenet, simonyan2014very, szegedy2015going,zagoruyko2016wide, he2016deep, xie2017aggregated, huang2017densely}. Usually these gains are the direct result of topology improvements and other mechanisms for efficient training, which is the case of the skip connections \cite{he2016deep}. The training procedure of all of the aforementioned networks was performed using backpropagation, which is the default standard for neural network training.


\section{Proposed Method}
\label{sec:proposed}
\begin{figure}[!t]
    \centering
    {\includegraphics[height=\textwidth,width=0.53\columnwidth]{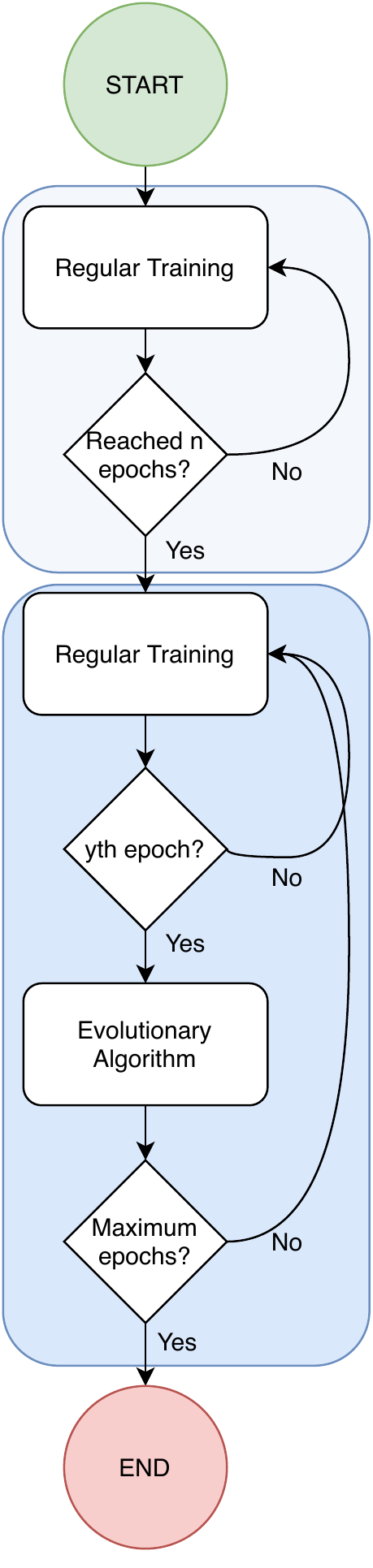}}
    \caption{Diagram of the training procedure. First, the regular training is performed $n$ epochs. After this, regular training and the evolutionary algorithm perform the training until reaching the limit of epochs, whereas, the evolutionary algorithm takes place every $y_{th}$ epoch.\label{fig:flow}}
\end{figure}

Inspired by transfer learning \cite{bengio2012deep}, we propose a hybrid method to complement backpropagation \cite{rumelhart1985learning} in the task of adjusting weights in CNNs training. The proposed method works by complementing the traditional training with an evolutionary algorithm, that evolves the weights of the fully connected output layer of a CNN. By doing so, the hybrid method assists the regular training in avoiding local minima, and, as it only evolves the weights of the last layer, it does so in useful time. The proposed method works by performing regular training (using only backpropagation) for a predefined number of epochs, $n$, and after those epochs are completed, the method continues to train until convergence, but every $y$ epochs, the evolutionary strategy takes place and evolves the weights of the last layer, in order to guide the train out of possible local minima. The reason to allow regular training for $n$ epochs initially is to allow the network to reach a point where its loss is low in a faster way than using both methods throughout the entire method.

The hybrid method is then composed of two steps: 1) regular training for $n$ epochs, 2) regular training until convergence, whilst having an evolutionary algorithm taking place every $y$ epochs, as represented in the diagram presented in Figure \ref{fig:flow}. The evolutionary algorithm developed is based on three components, the first component is the initial population, which is created using a heuristic initialization, whilst the other two components are focused on evolving the weights, being those the cross-overs and mutation, that replace the population by a new one at the end of a generation. Denote that elitism is also employed, where the best individual is kept without changes to the next generation, which forces both the algorithm to find weights that are closer to the best individuals and at the same time, ensures that in the worst case possible, where all generated weights are worse than the initial one, the performance of the CNN is not jeopardized, as if all are worse, elitism keeps the initial weights throughout the evolution. In the next sections, we explain in detail the three components mentioned.

\subsection{Initial Population}
\label{subsec:initpop}
To generate the initial population, we used a heuristic initialization, meaning that instead of randomly generating the initial population, we generate it based on a heuristic for the problem. The heuristic used is simply to mutate the weights coming from the regular training, from which we will generate $i-1$ individuals based on the initial weights, where the $i_{th}$ element is the initial weights without changes (use of elitism).

To generate the initial population, we then perform the following mutation, $i-1$ times, one for each new individual:
\begin{equation}
\label{equation:initpop}
    W = W_i + R*mag,\quad R\leftarrow \mathcal{N}(0,1)
\end{equation}

Where $W_i$ represent the weights obtained from the regular training, $R$ is a 1D array of pseudo-random generated numbers from a normal distribution, $mag$ is a real number, between $0$ and $1$ to smooth the mutations and $W$ is the resulting weights that represent the new individual.

\subsection{Evolution}
With the initial population, comprised of $i$ individuals, generated, the evolution takes place for $e$ generations. As aforementioned, each generation is composed of $i$ individuals, where the $i_{th}$ is the individual with the highest fitness from the previous generation. Moreover, one can think of the initial population as generation $0$, and the $i_{th}$ element of that generation as the original, unchanged, weights. To generate new individuals to compose the next generation, we implemented 3 types of mutations and 1 type of crossover, which are explained in detail in the next sections.

\subsubsection{Crossover}
To perform crossover in order to generate new individuals, we implemented a block-wise crossover, where $m$ individuals are selected to generate one offspring by contributing with a block of his own-weights, each. To select the $m$ individuals, we forced that only the top 30\% can be selected as parents and from that, we randomly select the $m$ parents, from the ordered set that make the top 30\%. To generate a new individual, first, we need to calculate the size of the blocks that each parent will contribute:
\begin{equation}
    k = l\ \mathbf{div}\ m 
\end{equation}
Where $l$ is the total size of the weights (length of the array), $m$ is the number of parents and $k$ is the result of the integer division of $l$ by $m$, representing the size of each block. With this, a new individual $A$, is generated by concatenating all blocks of the parents:
\begin{equation}
    A=[a_1,...a_m]
\end{equation}
Where $a$ represents a block from parent $1$ until parent $m$. Where each block $a$, from parent 1 until the $m_{th}$ parent, can be represented as:
\begin{equation}
    a_i = P_i[(i-1)*k:i*k], i=1,...,m-1
\end{equation}
Where $P$ is the parents array. Whereas the final block if there are more than one parent, $a_m$, is represented differently to accommodate odd sizes:
\begin{equation}
    a_m = P_m[(m-1)*k:l]
\end{equation}

\subsubsection{Mutation}
For weight mutation, which can be done after the crossover and applied to the new individual, or if crossover has not taken place, to one individual in the top 30\%, we implemented three methods. The first one is exactly the same as mentioned in Section \ref{subsec:initpop}, Equation \ref{equation:initpop}, where all the weights in the weight array suffer a small mutation. 

The second mutation is a block mutation, where the weight array of the individual being mutated is split into $b$ equal blocks, and one random block suffers the mutation:
\begin{equation}
    B = B + R*mag, \quad R\leftarrow \mathcal{U}(-B,B)
\end{equation}
Where $B$ represents the weights in the block, $mag$ is a real number, between $0$ and $1$ to smooth the mutations and $R$ is a 1D array of pseudo-random generated numbers from a uniform distribution, where each value is based on the value at the same position in the $B$ array. 

The last mutation implemented is a per value mutation, where for each element in the weights array, a pseudo-random number is generated, and if it is lower than a probability, $P_v$, the element suffers a mutation:
\begin{equation}
    v = v + r*mag, \quad r\leftarrow \mathcal{U}(-v,v)
\end{equation}
Where $v$ represents the element on a given position of the weights array, $r$ is a pseudo-random generated number using a uniform distribution, and $mag$ is a real number, between $0$ and $1$ to smooth the mutation.

Hence, the mutation algorithm is composed of the three aforementioned mutation protocols. It is important to denote that only one mutation can occur in a given individual, but as mentioned before, a mutation can take place in a newly generated individual by crossover.

\section{Experiments}
\label{sec:exp}
In order to evaluate the performance of the proposed method, we conducted an experiment using VGG16 \cite{simonyan2014very}, a well-known CNN. In the following sections, we explain the experimental setup, the dataset and the results obtained. 

\subsection{Experimental setup}
For this work, we used a computer with an NVidia GeForce GTX 1080 Ti, 16Gb of ram, 500GB SSD disk and an AMD Ryzen 7 2700 processor. Regarding the number of epochs for the training without the use of evolution, we set $n$ to 50, and $y$ to $10$, meaning that every $10$ epochs, starting at epoch number $50$, the evolution of the weights take place. We also set a maximum number of $100$ epochs for the entire training. Regarding the evolution, we set the number of individuals per generation, $i$, to 100 and the number of generations, $g$ to 5. Moreover, regarding the probability of crossover, we set it to $30\%$, and the probability of performing mutations to $50\%$, whereas, the probability of each mutation was set to: probability of mutation per value of $45\%$, probability of mutation per block of $45\%$ and probability of mutation of the entire array of $10\%$. Regarding the magnitude values, in Equation (1) we used a $mag=1$, and in Equation (6) and (7) $mag=0.05$.

We performed the experiment three times with the aforementioned setup, and to have a baseline, we also performed, three times, a simple regular train using only backpropagation for the entirety of the 100 epochs. Denote that we use the same seeds for the hybrid method and the regular method, meaning that three seeds were used, one for each experiment, where it was comprised of using the hybrid method and the regular method. 

\subsection{CIFAR-10}
CIFAR-10 is a dataset composed of $10$ classes with a total of $60,000$ RGB images with a size of $32$ by $32$ pixels \cite{cifar10}. It was created by labelling a subset of the $80$ Million Tiny Images Dataset. The official dataset is split into five training batches of $10$ thousand images each and a batch of $10$ thousand images for testing. The $10$ classes are: airplane, automobile, bird, cat, deer, dog, frog, horse, ship and truck. As the dataset was manually labelled, the authors took the precaution of ensuring that the classes are mutually exclusive, meaning that there is no overlap between any of two classes. Denote that the classes are balanced, each one having $6000$ images and, the evaluation metric used is accuracy. For the experiments conducted, we used $40,000$ images for the train, $10,000$ images for validation and $10,000$ images for the test set, with stratified and balanced splits.

\subsection{Results and Discussion}
In table \ref{tab:label_test}, we show the accuracy and standard deviation for each epoch where the evolution could take place (epoch 50 and 10 epoch after), for both the hybrid method and the regular training. For the hybrid method, we also state the performance at the end of each generation. From this, we can see that the hybrid method is capable of further improving the weights of the regular training, meaning that it achieves higher performance, every time the evolution takes place, than using only the regular method for training. Regarding the performance of the CNN in the validation set and in the test set, after completing the process of training, and the time cost of training both the hybrid method and the regular method (baseline), can be seen in Table \ref{tab:final-results}. In this, it is possible to see that the hybrid method is capable of complementing backpropagation and achieving higher performances than the regular method in both the validation and test sets. However, this accuracy improvement in performances comes at a higher time cost.

\begin{table*}[!t]
\resizebox{\textwidth}{!}{  
\begin{tabular}{c|llll|l|l}
\toprule
\multirow{2}{*}{Epoch} & \multicolumn{5}{c|}{Hybrid Method Accuracy/Generation (\%) }                                                                                            & Regular Method  \\
                       & \multicolumn{1}{c}{1} & \multicolumn{1}{c}{2} & \multicolumn{1}{c}{3} & \multicolumn{1}{c}{4} & \multicolumn{1}{c|}{5} & \multicolumn{1}{c}{Accuracy (\%)}                       \\ \midrule
50                     & $88.09 \pm 0.25$          & $88.12 \pm 0.23$          & $88.18 \pm 0.21$          & $88.18 \pm 0.21$          & \boldmath$88.21 \pm 0.19$          & $87.88 \pm 0.32$                                 \\
60                     & $88.37 \pm 0.15$          & $88.40 \pm 0.16$          & $88.42 \pm 0.17$          & $88.42 \pm 0.17$          & \boldmath$88.45 \pm 0.17$          & $88.24 \pm 0.07$                                 \\
70                     & $88.20 \pm 0.48$          & $88.26 \pm 0.47$         & $88.29 \pm 0.48$          & $88.32 \pm 0.50$          & \boldmath$88.34 \pm 0.51$          & $88.08 \pm 0.39$                                 \\
80                     & $88.52 \pm 0.11$          & $88.56 \pm 0.13$          & $88.58 \pm 0.13$          & $88.60 \pm 0.13$          & \boldmath$88.61 \pm 0.13$          & $88.43 \pm 0.11$                                 \\
90                     & $88.47 \pm 0.13$          & $88.48 \pm 0.15$          & $88.51 \pm 0.12$          & $88.54 \pm 0.10$          & \boldmath$88.54 \pm 0.11$          & $88.37 \pm 0.13$                                 \\
100                    & $88.63 \pm 0.13$          & $88.65 \pm 0.13$          & $88.67 \pm 0.12$          & $88.70 \pm 0.11$          & \boldmath$88.71 \pm 0.12$          & $88.00 \pm 0.64$  \\ \bottomrule                        
\end{tabular}
}
\caption{Mean validation accuracy and standard deviation in every 10 epochs after epoch 50 for both the regular training alone, and for the proposed hybrid method, where the accuracy is shown for every generation, denoting that the fifth generation is the end of the epoch.}
  \label{tab:label_test}
\end{table*}

\begin{table}[!t]
\centering
\caption{Mean accuracy and standard deviation in the validation and test set and the time spent for training and testing both the Hybrid Method and the Regular Method (baseline). \label{tab:final-results}}
\resizebox{\columnwidth}{!}{  
\begin{tabular}{l|cc|c}
\toprule
               & \begin{tabular}[c]{@{}c@{}}Validation Accuracy\\ (\%)\end{tabular} & \multicolumn{1}{c|}{\begin{tabular}[c]{@{}c@{}}Test Accuracy\\ (\%)\end{tabular}} & \multicolumn{1}{c}{\begin{tabular}[c]{@{}c@{}}Time Cost\\ (s)\end{tabular}} \\ \midrule
Hybrid Method  & \boldmath$88.77 \pm 0.40$                                     & \boldmath$88.41 \pm 0.43$                                                    & 9780.98                                                                     \\
Regular Method & $88.45 \pm 0.15$                                     & $87.80 \pm 0.04$                                                    & \textbf{2363.78} \\ \bottomrule                                                                    
\end{tabular}
}
\end{table}

\section{Conclusion}
\label{sec:conc}
Deep learning has extended the use of Artificial Intelligence to virtually every field of research, due to its capability of learning in an automated way, without requiring extensive human-extracted features. To train CNNs, backpropagation is the go-to algorithm. It uses gradients to adjust the weights, resulting in trained CNNs with higher performances than randomly generated weights. Many proposals fine-tune the weights to achieve even better results, which is the case of transfer learning, where one CNN is trained in a given task, and then the weights are re-used to train a CNN with the same topology (or close, as the last layer is normally changed due to classification neurons) in other tasks. This inspired us to create a hybrid method that uses both the traditional, regular training with backpropagation and evolutionary strategies to fine-tune the weights of the last fully connected layer. The proposed method is capable of further improving the results of a CNN and can be used to fine-tune CNNs. This work shows that there is potential in using hybrid methods, since in our experiments the hybrid method was capable of achieving better results than just using a regular method (only backpropagation). The down-side of this is that evolutionary strategies often require great computational power, as they usually perform unguided searches throughout the search space and require many individuals to find good candidates. However, this presents research opportunities in the field of evolving only a set of weights as candidates, which can be further extended to new problems and using evolutionary mechanisms that are known to improve the performance of such systems. Therefore, this method can be used in two ways: 1) as presented here, where a CNN can be trained from scratch using the hybrid method, or 2) just fine-tuning a neural network that is already trained, to find weights that can achieve better performances. 

{\footnotesize
\bibliographystyle{ieee_fullname}
\bibliography{egbib}
}

\end{document}